\documentclass[letterpaper]{article} 
\usepackage{aaai2027}  
\usepackage[hyphens]{url}  
\usepackage{graphicx} 
\usepackage{natbib}  
\usepackage{caption} 
\usepackage{algorithm}
\usepackage{algorithmic}
\usepackage{amsmath, amssymb, amsthm}
\usepackage{mathtools}
\usepackage{bm}
\usepackage{booktabs}
\usepackage{cleveref}
\usepackage{xcolor}
\usepackage{thmtools}
\usepackage{mdframed}
\usepackage{microtype}
\usepackage{adjustbox}

\usepackage{siunitx}

\theoremstyle{definition}

\usepackage{amsmath}

\newcommand{\best}[2]{%
    \ensuremath{\boldsymbol{#1{\scriptstyle\pm#2}}}%
}
\newcommand{\secondbest}[2]{%
    \ensuremath{\underline{#1{\scriptstyle\pm#2}}}%
}
\newcommand{\R}{\mathbb{R}}
\newcommand{\E}{\mathbb{E}}

\newcommand{\method}{EulerLoRA}
\newcommand{\1}{\mathbf{1}}
\newcommand{\softmax}{\operatorname{softmax}}
\usepackage{newfloat}
\usepackage{listings}
\DeclareCaptionStyle{ruled}{labelfont=normalfont,labelsep=colon,strut=off} 
\floatstyle{ruled}
\newfloat{listing}{tb}{lst}{}
\floatname{listing}{Listing}

\usepackage{booktabs}

\nocopyright 

\title{EulerLoRA: Rank-Driven Jump Dynamics for Calibrated Parameter-Efficient Fine-Tuning
}
\author{
    Srinivas Anumasa, Dianbo Liu\textsuperscript{\rm 1}
}
\affiliations{
    \textsuperscript{\rm 1}National University of Singapore\\

}

\begin{document}

\maketitle


\begin{abstract}
Low-Rank Adaptation (LoRA) enables parameter-efficient fine-tuning,
but standard LoRA produces a single deterministic model and does not
directly support predictive uncertainty estimation. We introduce
EulerLoRA, a stochastic extension of LoRA that generates multiple
predictive trajectories by sampling structured variations along the
rank-one components of shared low-rank adapters, while preserving the
deterministic LoRA transformation in expectation. We evaluate EulerLoRA with vision transformers on CIFAR-10,
CIFAR-100, and HAM10000, together with out-of-distribution detection
on SVHN. Across these benchmarks, EulerLoRA achieves comparable or
improved performance 
relative to strong LoRA-Ensemble baselines. Using two rank-$20$
adapters, EulerLoRA requires approximately $3$ million trainable
adapter parameters, compared with about $10$ million for a
rank-$8$, 16-adapter LoRA-Ensemble, corresponding to roughly $69\%$
fewer trainable parameters. These results show that useful predictive
diversity can be obtained from a small number of shared adapters.
\end{abstract}

\section{Introduction}

Large pretrained models
\cite{devlin2019bert,brown2020language,dosovitskiy2021image}
have achieved strong transfer performance across a wide range of
downstream tasks
\cite{raffel2020exploring,radford2021learning}. However, adapting all pretrained parameters
for every task incurs substantial computational and storage costs.
Parameter-efficient fine-tuning methods address this limitation by updating
only a small subset of parameters while keeping most of the pretrained model
frozen. Among these methods, LoRA \cite{hu2022lora} and its variants
\cite{zhang2023adalora,valipour2023dylora,liu2024dora}
have become prominent approaches to parameter-efficient adaptation. For a pretrained weight
matrix $\mathbf{W}_0$, LoRA learns a low-rank correction
$\Delta\mathbf{W}=\mathbf{B}\mathbf{A}$, where the rank of the update is much
smaller than the input and output dimensions of $\mathbf{W}_0$. This
parameterization substantially reduces the number of trainable parameters. 
Originally introduced for large language models, LoRA has since been
adopted beyond language modelling, including for vision transformers
\cite{muehlematter2026loraensemble} and generative diffusion models
\cite{kasymov2024autolora}.

Despite its strong predictive performance, accuracy alone is insufficient
for assessing the reliability of a fine-tuned model. Modern neural networks
can be miscalibrated, assigning confidence scores that do not accurately
reflect their probability of being correct
\cite{guo2017calibration}, and their uncertainty estimates can degrade
further under distribution shift
\cite{ovadia2019trust}. These concerns remain relevant in
parameter-efficient fine-tuning. Standard LoRA learns a single pair of
low-rank matrices for each adapted weight, yielding a deterministic update
$\Delta W = BA$ after training \cite{hu2022lora}. Although the resulting
model can produce confidence scores such as softmax probabilities or
predictive entropy, it does not by itself define a distribution over the
adapter parameters or the corresponding predictive functions. Consequently,
standard LoRA provides no direct mechanism for representing epistemic
uncertainty through posterior marginalization or model averaging. This
limitation has motivated recent Bayesian and ensemble-based extensions of
LoRA \cite{yang2024bayesian,wang2024blob,
muehlematter2026loraensemble}.
\begin{figure*}[t]
    \centering
    \includegraphics[width=\textwidth]{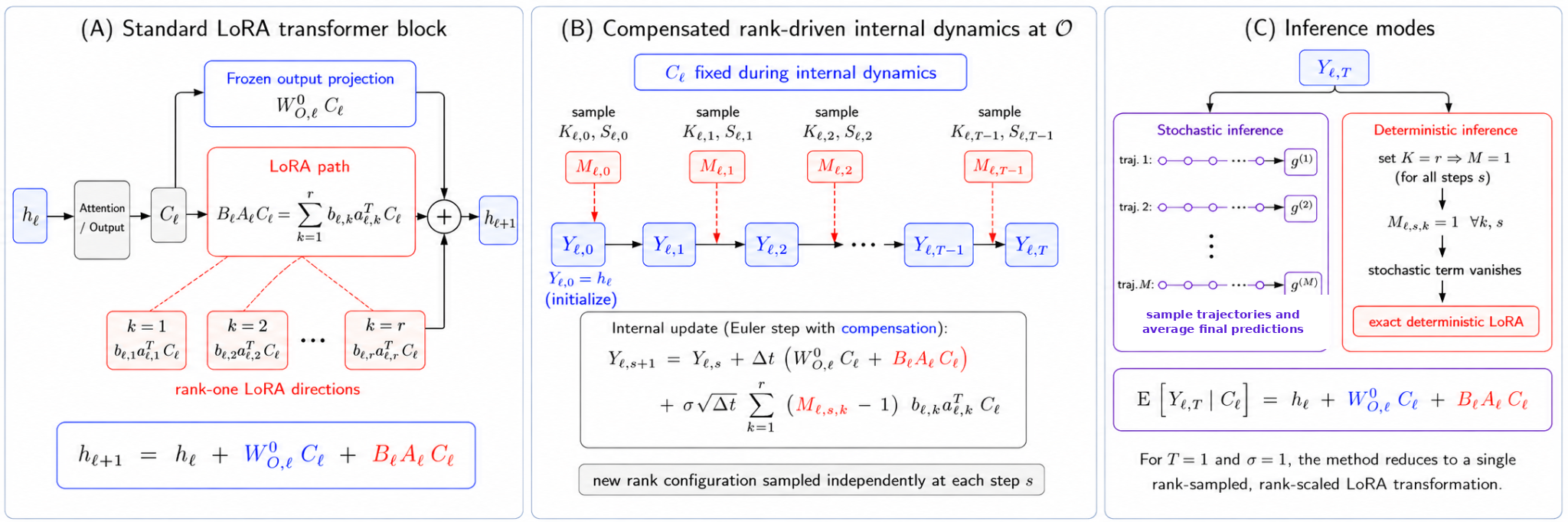}
  \caption{
Overview of the proposed compensated rank-driven dynamics for LoRA.
\textbf{(A)} Standard LoRA transformer block, where the output projection
is decomposed into a frozen pretrained path and a low-rank LoRA path
consisting of learned rank-one directions.
\textbf{(B)} Compensated internal dynamics at the output projection \(O\).
At each internal step \(s\), a rank configuration
\((K_{\ell,s},S_{\ell,s})\) is sampled, defining the rank-scaled coefficients $M_{\ell,s,k}$.  While \(C_\ell\) remains fixed, the auxiliary state evolves over \(T\)
finite Euler steps using the deterministic LoRA update together with
centered rank-dependent fluctuations.
\textbf{(C)} In stochastic inference, multiple sampled trajectories are
aggregated for uncertainty estimation. In deterministic inference,
setting \(K_{\ell,s}=r\) at every step gives \(M_{\ell,s,k}=1\), causing the
stochastic term to vanish and recovering exact deterministic LoRA.
Conditional on $C_\ell$, the expected terminal state recovers the
deterministic output-projection LoRA transformation.
}
    \label{fig:lora_jump_dynamics}
\end{figure*}
Recent work has begun to address this limitation by introducing uncertainty
directly into the low-dimensional adapter space. Laplace-LoRA applies a
post-hoc Laplace approximation to the learned LoRA parameters
\cite{yang2024bayesian}, while BLoB learns a variational distribution over
the low-rank factors during fine-tuning \cite{wang2024blob}. An alternative
direction is LoRA-Ensemble, which shares a common pretrained backbone but
learns a separate LoRA adapter for each ensemble member
\cite{muehlematter2026loraensemble}. These approaches demonstrate that useful
uncertainty estimates can be obtained without fine-tuning the full model.
However, Bayesian formulations require additional posterior
approximations or distributional parameters, whereas ensemble-based
methods increase the number of learned adapter parameters with the
number of ensemble members. This motivates the following question:
can multiple useful predictive realizations be generated from the same
learned low-rank adaptation space, without training a separate adapter
for every prediction? Rather than learning an independent adapter for
each predictive member, we seek to introduce structured stochasticity
along the rank-one directions of a shared LoRA update, while preserving
the underlying deterministic LoRA transformation in expectation.

To realize this objective, we draw on the dynamical-systems
interpretation of deep neural networks
\cite{haber2018stable,lu2018beyond}. Under this view, network depth
acts as discrete time, while the hidden representations trace the
state trajectory of an underlying dynamical system. This connection
is particularly transparent in residual networks, whose layerwise
update takes the form
\begin{equation}
    h_{\ell+1}
    =
    h_{\ell}+F_{\ell}(h_{\ell}).
\end{equation}
The identity path preserves the current state, while the residual
branch supplies an additive increment. The resulting update has the
form of a forward Euler discretization of an ordinary differential
equation, with the layer index playing the role of time. This
connection motivated continuous-depth architectures such as Neural
Ordinary Differential Equations \cite{chen2018neural}. Extensions of
this viewpoint have interpreted residual networks with random
increments as discretizations of stochastic differential systems
\cite{sun2018stochastic,liu2020noise}, while neural jump models
incorporate discontinuous state changes into the underlying dynamics
\cite{jia2019neuraljump}. Inspired by these connections, we construct
finite-step stochastic dynamics in which each sampled rank
configuration determines a structured update along the learned
rank-one directions of LoRA. Different realizations therefore produce
multiple predictive trajectories while sharing the same low-rank
adapter parameters.

Building on this viewpoint, we introduce \emph{EulerLoRA}, a stochastic
formulation of LoRA based on finite-step rank-driven dynamics. For clarity,
we describe the construction at the output projection of a self-attention
module. Let $\mathbf{C}_{\ell}$ denote the attention representation entering
this projection. The corresponding LoRA correction can be decomposed into
$r$ rank-one components:
\begin{equation}
    \mathbf{B}_{\ell}\mathbf{A}_{\ell}\mathbf{C}_{\ell}
    =
    \sum_{k=1}^{r}
    \mathbf{b}_{\ell k}
    \mathbf{a}_{\ell k}^{\top}
    \mathbf{C}_{\ell}.
\end{equation}
Each rank-one component therefore defines a learned direction along which
the LoRA correction can evolve. Standard LoRA corresponds to the
deterministic transformation in which every component contributes with a
unit coefficient. EulerLoRA instead samples rank configurations over a
finite number of internal steps and uses them to generate stochastic
fluctuations along these shared rank-one directions.

A direct accumulation of rank-sampled updates would cause the expected
LoRA contribution to depend on the number of internal steps. We avoid this
by separating the deterministic LoRA transformation from the centered
stochastic fluctuations induced by rank sampling. The deterministic term
is accumulated with Euler step-size scaling, whereas only the zero-mean
rank fluctuations receive stochastic scaling. Consequently, the expected
terminal state exactly recovers the standard LoRA transformation,
independently of the number of internal steps. Setting every rank
coefficient to one removes the stochastic term and recovers deterministic
LoRA exactly.

Different realizations of the rank configurations generate different
predictive trajectories while sharing the same pretrained backbone and
learned low-rank matrices. During training, this stochastic rank switching
exposes the model to a family of structured perturbations within the
learned adaptation space. During inference, multiple realizations form an
implicit Monte Carlo ensemble, enabling predictive uncertainty estimation
without learning a separate adapter for every predictive sample. The
number of predictive realizations can therefore be increased independently
of the number of learned adapters. EulerLoRA also supports deterministic
inference by retaining all rank components, allowing the same trained
model to be used either as a deterministic predictor or as a stochastic
uncertainty model.

We evaluate EulerLoRA using pretrained vision transformers on CIFAR-10,
CIFAR-100, and HAM10000, together with out-of-distribution detection on
SVHN. We assess classification accuracy and macro-F1, calibration through
expected calibration error, proper scoring rules through negative
log-likelihood and Brier score, and out-of-distribution detection using
maximum softmax probability. The results show
that deterministic and stochastic inference provide complementary
benefits. Deterministic inference generally provides the strongest
predictive accuracy and proper scoring performance, whereas stochastic
inference can improve calibration and out-of-distribution detection.
Using only two rank-$20$ adapters, EulerLoRA remains competitive with, and
on several metrics improves upon, a substantially larger rank-$8$,
16-adapter LoRA-Ensemble. Figure \ref{fig:lora_jump_dynamics} gives an overview our proposed method. 

\paragraph{Contributions.}
Our main contributions are summarized as follows:
\begin{itemize}
    \item We introduce EulerLoRA, a stochastic formulation of LoRA in
    which the rank-one components of the low-rank correction define
    structured directions for finite-step rank-driven dynamics.

    \item We derive a mean-preserving construction that separates the
    deterministic LoRA transformation from zero-mean stochastic rank
    fluctuations. The expected terminal state exactly recovers standard
    LoRA independently of the number of internal steps, while the
    deterministic limit recovers the ordinary LoRA transformation.

    \item We generate multiple predictive trajectories from shared LoRA
    parameters, thereby decoupling the number of Monte Carlo predictions
    from the number of independently trained adapters.

    \item We evaluate the proposed method across three image-classification
    datasets and an out-of-distribution detection benchmark, demonstrating
    complementary benefits of deterministic and stochastic inference for
    predictive performance, calibration, and uncertainty estimation.
\end{itemize}

\section{Compensated Rank-Driven Jump Dynamics}

This section develops EulerLoRA from the residual structure of a
LoRA-adapted transformer block. We first introduce an auxiliary
internal state that evolves only along the additive output path, thereby
avoiding repeated application of the frozen pretrained transformation.
We then decompose the LoRA correction into rank-one directions and
define expectation-preserving stochastic rank configurations. Finally,
we show why a direct stochastic formulation produces an undesirable
dependence on the number of internal steps, derive the compensated
dynamics, and establish their mean-preserving and deterministic limits.
\paragraph{From residual adaptation to internal dynamics.}
Consider the output of the $\ell$-th transformer block equipped with
LoRA. Let $\mathbf{C}_{\ell}$ denote the attention representation
entering the output projection. The resulting residual update is
\begin{equation}
\mathbf{h}_{\ell+1}
=
\mathbf{h}_{\ell}
+
\mathbf{W}^{0}_{O,\ell}\mathbf{C}_{\ell}
+
\mathbf{B}_{\ell}\mathbf{A}_{\ell}\mathbf{C}_{\ell},
\label{eq:lora_block}
\end{equation}
where $\mathbf{W}^{0}_{O,\ell}$ is the frozen pretrained output
projection and
$\mathbf{B}_{\ell}\mathbf{A}_{\ell}\mathbf{C}_{\ell}$
is the learned LoRA correction.

Equation~\eqref{eq:lora_block} has the additive structure of a residual
update: the current representation $\mathbf{h}_{\ell}$ is preserved by
the identity path, while the pretrained projection and the LoRA
correction form an additive increment. As discussed in the
introduction, residual updates admit a forward-Euler interpretation,
with network depth corresponding to discrete time. This observation
suggests that the additive transformation within a transformer block
can itself be refined over a finite number of internal steps, without
introducing additional trainable parameters.

A direct temporalization of the complete transformer block, however,
would be inappropriate for parameter-efficient adaptation. Repeatedly
evolving the hidden state would also repeatedly apply the frozen
pretrained transformation
$\mathbf{W}^{0}_{O,\ell}\mathbf{C}_{\ell}$.
Consequently, when $T>1$, the resulting architecture would no longer
recover the original pretrained transformer even when the LoRA
parameters are initialized to zero. We therefore do not evolve the
complete hidden representation. Instead, we keep
$\mathbf{C}_{\ell}$ fixed and introduce an auxiliary state
$\mathbf{Y}_{\ell,s}$ that evolves only along the additive output path,
with
\begin{equation}
\mathbf{Y}_{\ell,0}=\mathbf{h}_{\ell}.
\end{equation}
After $T$ internal steps, the terminal state
$\mathbf{Y}_{\ell,T}$ is passed to the next transformer block.

\paragraph{Rank-one directions of the LoRA update.}
The low-rank correction in Eq.~\eqref{eq:lora_block} can be decomposed
as
\begin{equation}
\mathbf{B}_{\ell}\mathbf{A}_{\ell}\mathbf{C}_{\ell}
=
\sum_{k=1}^{r}
\mathbf{b}_{\ell,k}
\mathbf{a}_{\ell,k}^{\top}
\mathbf{C}_{\ell},
\label{eq:rank_one_decomp}
\end{equation}
where $\mathbf{b}_{\ell,k}$ is the $k$-th column of
$\mathbf{B}_{\ell}$ and $\mathbf{a}_{\ell,k}^{\top}$ is the $k$-th
row of $\mathbf{A}_{\ell}$. Each term in
Eq.~\eqref{eq:rank_one_decomp} defines a learned rank-one direction
along which the additive correction state can evolve.

Standard LoRA corresponds to the deterministic case in which every
rank-one direction contributes with coefficient one. To obtain
multiple predictive trajectories from the same learned adapter, we
allow the participating directions and their coefficients to vary
across the internal steps. Since these changes occur at fixed discrete
steps and are determined by sampled rank configurations, we interpret
the resulting construction as finite-step rank-driven dynamics, with
the sampled configuration acting as the mark of each update.

\paragraph{Stochastic rank configurations.}
We divide the unit interval into $T$ internal steps and set
$\Delta t=1/T$. At step $s$, we sample an active rank
\begin{equation}
K_{\ell,s}\in\{1,\ldots,r\},
\end{equation}
followed by a subset
\begin{equation}
\mathcal{S}_{\ell,s}
\subseteq
\{1,\ldots,r\},
\qquad
|\mathcal{S}_{\ell,s}|=K_{\ell,s},
\end{equation}
chosen uniformly among all subsets of size $K_{\ell,s}$. The
coefficient of the $k$-th rank-one direction is
\begin{equation}
M_{\ell,s,k}
=
\frac{r}{K_{\ell,s}}
\mathbf{1}\{k\in\mathcal{S}_{\ell,s}\}.
\label{eq:rank_mask}
\end{equation}

Conditioned on $K_{\ell,s}$, every rank component has probability
$K_{\ell,s}/r$ of being selected. Therefore,
\begin{equation}
\mathbb{E}
\left[
M_{\ell,s,k}\mid K_{\ell,s}
\right]
=
\frac{r}{K_{\ell,s}}
\frac{K_{\ell,s}}{r}
=
1.
\label{eq:mask_expectation}
\end{equation}
Thus, although individual rank configurations retain only a subset of
the rank-one directions, the scaled coefficient preserves each
component in expectation.

\paragraph{Why compensation is required.}
Given the sampled coefficients in Eq.~\eqref{eq:rank_mask}, a direct
variance-scaled evolution of the auxiliary state would take the form
\begin{align}
\mathbf{Y}_{\ell,s+1}
={}&
\mathbf{Y}_{\ell,s}
+
\Delta t\,
\mathbf{W}^{0}_{O,\ell}\mathbf{C}_{\ell}
\nonumber\\
&+
\sigma\sqrt{\Delta t}
\sum_{k=1}^{r}
M_{\ell,s,k}
\mathbf{b}_{\ell,k}
\mathbf{a}_{\ell,k}^{\top}
\mathbf{C}_{\ell},
\label{eq:direct_rank_dynamics}
\end{align}
where $\sigma\geq 0$ controls the magnitude of the stochastic
rank-dependent update.

Although the factor $\sqrt{\Delta t}$ is appropriate for controlling
the variance accumulated over multiple internal steps, the sampled
rank term in Eq.~\eqref{eq:direct_rank_dynamics} is not zero mean.
Conditioned on $\mathbf{C}_{\ell}$ and $K_{\ell,s}$, we have
\begin{align}
&\mathbb{E}\!\left[
\sum_{k=1}^{r}
M_{\ell,s,k}
\mathbf{b}_{\ell,k}
\mathbf{a}_{\ell,k}^{\top}
\mathbf{C}_{\ell}
\,\middle|\,
\mathbf{C}_{\ell},K_{\ell,s}
\right]
\nonumber\\
&\qquad =
\sum_{k=1}^{r}
\mathbb{E}\!\left[
M_{\ell,s,k}\mid K_{\ell,s}
\right]
\mathbf{b}_{\ell,k}
\mathbf{a}_{\ell,k}^{\top}
\mathbf{C}_{\ell}
\nonumber\\
&\qquad =
\mathbf{B}_{\ell}\mathbf{A}_{\ell}\mathbf{C}_{\ell}.
\label{eq:sampled_update_expectation}
\end{align}
Thus, the rank-sampled term contains both the predictable LoRA
transformation and a random deviation around it.

If Eq.~\eqref{eq:direct_rank_dynamics} is applied for $T$ internal
steps, its expected LoRA contribution becomes
\begin{equation}
T\sigma\sqrt{\Delta t}\,
\mathbf{B}_{\ell}\mathbf{A}_{\ell}\mathbf{C}_{\ell}
=
\sigma\sqrt{T}\,
\mathbf{B}_{\ell}\mathbf{A}_{\ell}\mathbf{C}_{\ell},
\end{equation}
because $\Delta t=1/T$. The expected transformation would therefore
change with the number of internal steps instead of recovering a fixed
LoRA correction. This occurs because the predictable and stochastic
parts of the sampled rank update are being assigned the same
$\sqrt{\Delta t}$ scaling.

\paragraph{Compensated rank-driven dynamics.}
To separate the predictable LoRA transformation from the stochastic
rank fluctuations, we decompose each sampled coefficient as
\begin{equation}
M_{\ell,s,k}
=
1+\left(M_{\ell,s,k}-1\right).
\label{eq:mask_decomposition}
\end{equation}
By Eq.~\eqref{eq:mask_expectation},
\begin{equation}
\mathbb{E}
\left[
M_{\ell,s,k}-1
\mid K_{\ell,s}
\right]
=0.
\end{equation}
The term $M_{\ell,s,k}-1$ therefore represents the centered deviation
of the sampled coefficient from its conditional mean.

We assign the deterministic pretrained and LoRA transformations the
Euler scaling $\Delta t$, so that they accumulate exactly once over
the unit interval. Only the centered rank-dependent fluctuation is
assigned the stochastic scaling $\sqrt{\Delta t}$. The resulting
compensated dynamics are
\begin{align}
\mathbf{Y}_{\ell,s+1}
={}&
\mathbf{Y}_{\ell,s}
+
\Delta t
\left(
\mathbf{W}^{0}_{O,\ell}\mathbf{C}_{\ell}
+
\mathbf{B}_{\ell}\mathbf{A}_{\ell}\mathbf{C}_{\ell}
\right)
\nonumber\\
&+
\sigma\sqrt{\Delta t}
\sum_{k=1}^{r}
\left(M_{\ell,s,k}-1\right)
\mathbf{b}_{\ell,k}
\mathbf{a}_{\ell,k}^{\top}
\mathbf{C}_{\ell}.
\label{eq:compensated_dynamics}
\end{align}

We refer to Eq.~\eqref{eq:compensated_dynamics} as compensated because
the conditional mean of the rank-sampled update is removed from the
stochastic term and incorporated explicitly into the deterministic
increment. Consequently, the stochastic component describes only
zero-mean fluctuations around the ordinary LoRA transformation.

\paragraph{Terminal state and mean preservation.}
Summing Eq.~\eqref{eq:compensated_dynamics} over
$s=0,\ldots,T-1$ and using
$\mathbf{Y}_{\ell,0}=\mathbf{h}_{\ell}$ and
$\Delta t=1/T$ gives
\begin{align}
\mathbf{Y}_{\ell,T}
={}&
\mathbf{h}_{\ell}
+
\mathbf{W}^{0}_{O,\ell}\mathbf{C}_{\ell}
+
\mathbf{B}_{\ell}\mathbf{A}_{\ell}\mathbf{C}_{\ell}
\nonumber\\
&+
\frac{\sigma}{\sqrt{T}}
\sum_{s=0}^{T-1}
\sum_{k=1}^{r}
\left(M_{\ell,s,k}-1\right)
\mathbf{b}_{\ell,k}
\mathbf{a}_{\ell,k}^{\top}
\mathbf{C}_{\ell}.
\label{eq:terminal_state}
\end{align}

Since each centered coefficient has zero conditional mean,
\begin{equation}
\mathbb{E}\!\left[
\mathbf{Y}_{\ell,T}
\mid
\mathbf{C}_{\ell}
\right]
=
\mathbf{h}_{\ell}
+
\mathbf{W}^{0}_{O,\ell}\mathbf{C}_{\ell}
+
\mathbf{B}_{\ell}\mathbf{A}_{\ell}\mathbf{C}_{\ell}.
\label{eq:terminal_expectation}
\end{equation}
The expected terminal state therefore exactly recovers the standard
deterministic LoRA block and is independent of the number of internal
steps $T$. Individual trajectories, however, differ through the
accumulated centered fluctuations along the learned rank-one
directions.

\paragraph{Special cases and interpretation.}
When $K_{\ell,s}=r$, the sampled subset contains all rank components,
and
\begin{equation}
M_{\ell,s,k}=1
\qquad
\text{for every } k.
\end{equation}
The stochastic term in Eq.~\eqref{eq:compensated_dynamics} then
vanishes, and the terminal state reduces exactly to the deterministic
LoRA transformation. This provides the deterministic inference mode
used in our experiments.

For $T=1$ and $\sigma=1$, Eq.~\eqref{eq:terminal_state} becomes
\begin{align}
\mathbf{Y}_{\ell,1}
={}&
\mathbf{h}_{\ell}
+
\mathbf{W}^{0}_{O,\ell}\mathbf{C}_{\ell}
\nonumber\\
&+
\sum_{k=1}^{r}
M_{\ell,0,k}
\mathbf{b}_{\ell,k}
\mathbf{a}_{\ell,k}^{\top}
\mathbf{C}_{\ell},
\end{align}
which is a single rank-sampled and rank-scaled LoRA transformation.
For $T>1$, the terminal coefficient of a rank-one direction is formed
by accumulating centered contributions across internal steps.
Accordingly, the resulting coefficient need not be binary or
nonnegative. The proposed construction is therefore not equivalent to
repeated conventional rank dropout; it produces stochastic deviations
around the deterministic LoRA transformation through finite-step
rank-driven dynamics.

\section{Experiments}
\begin{table*}[t]
    \centering
    \caption{Performance comparison on CIFAR-10. Here, R-$r$ denotes
    LoRA rank $r$, and Ad-$m$ denotes $m$ learned adapters. Results for
    the standard baselines and LoRA-Ensemble (R-8, Ad-16) are taken
    from the original LoRA-Ensemble paper. The reproduced rank-$20$
    baselines and EulerLoRA results are reported as mean $\pm$ standard
    deviation over five seeds. EulerLoRA uses $T=2$ dynamics steps.
    Stochastic inference uses four Monte Carlo samples per adapter,
    whereas deterministic inference uses a single evaluation pass. Both use same trained
EulerLoRA checkpoints and differ only in the inference mode.}
    \label{tab:cifar10_results}
    \resizebox{\linewidth}{!}{
    \begin{tabular}{lccccc}
        \toprule
        \textbf{Method}
        & \textbf{Accuracy ($\uparrow$)}
        & \textbf{F1 ($\uparrow$)}
        & \textbf{ECE ($\downarrow$)}
        & \textbf{NLL ($\downarrow$)}
        & \textbf{Brier ($\downarrow$)} \\
        \midrule

        Single Network
        & $92.8 \pm 0.1$
        & $92.8 \pm 0.1$
        & $0.051 \pm 0.001$
        & $0.333 \pm 0.003$
        & $0.120 \pm 0.002$ \\

        Single Net w/ LoRA
        & $94.5 \pm 0.0$
        & $94.5 \pm 0.0$
        & $0.009 \pm 0.001$
        & $0.163 \pm 0.002$
        & $0.082 \pm 0.001$ \\

        LoRA (R-20, Ad-1, Reprod.)
        & $94.66 \pm 0.14$
        & $94.66 \pm 0.14$
        & $0.022 \pm 0.001$
        & $0.173 \pm 0.004$
        & $0.083 \pm 0.002$ \\

        MC Dropout
        & $92.9 \pm 0.2$
        & $92.9 \pm 0.2$
        & $0.023 \pm 0.002$
        & $0.260 \pm 0.005$
        & $0.110 \pm 0.003$ \\

        Snapshot Ensemble
        & $93.1 \pm 0.1$
        & $93.1 \pm 0.1$
        & $0.037 \pm 0.002$
        & $1.062 \pm 0.021$
        & $0.510 \pm 0.008$ \\

        Batch-Ensemble
        & $88.5 \pm 0.1$
        & $88.5 \pm 0.1$
        & $0.048 \pm 0.001$
        & $0.347 \pm 0.001$
        & $0.172 \pm 0.000$ \\

        Explicit Ensemble
        & $94.1 \pm 0.1$
        & $94.1 \pm 0.1$
        & $0.031 \pm 0.001$
        & $0.181 \pm 0.002$
        & $0.087 \pm 0.001$ \\

        LoRA-Ensemble (R-8, Ad-16)
        & $95.9 \pm 0.1$
        & $95.9 \pm 0.1$
        & \best{0.003}{0.001}
        & $0.128 \pm 0.001$
        & \secondbest{0.064}{0.000} \\

        LoRA-Ensemble (R-20, Ad-2)
        & $95.43 \pm 0.05$
        & $95.43 \pm 0.05$
        & $0.007 \pm 0.001$
        & $0.138 \pm 0.002$
        & $0.069 \pm 0.001$ \\

        \midrule

        \textbf{EulerLoRA} (R-20, Ad-1, Stoch.)
        & $95.44 \pm 0.17$
        & $95.44 \pm 0.17$
        & $0.007 \pm 0.001$
        & $0.135 \pm 0.003$
        & $0.069 \pm 0.002$ \\

        \textbf{EulerLoRA} (R-20, Ad-1, Det.)
        & $94.69 \pm 0.24$
        & $94.69 \pm 0.24$
        & $0.009 \pm 0.002$
        & $0.156 \pm 0.003$
        & $0.079 \pm 0.002$ \\

        \textbf{EulerLoRA} (R-20, Ad-2, Stoch.)
        & \secondbest{95.92}{0.08}
        & \secondbest{95.93}{0.08}
        & $0.016 \pm 0.001$
        & \secondbest{0.126}{0.002}
        & \secondbest{0.064}{0.001} \\

        \textbf{EulerLoRA} (R-20, Ad-2, Det.)
        & \best{96.10}{0.05}
        & \best{96.10}{0.05}
        & \secondbest{0.004}{0.000}
        & \best{0.116}{0.001}
        & \best{0.059}{0.001} \\

        \bottomrule
    \end{tabular}
    }
\end{table*}

\begin{table}[t]
    \centering
    \caption{OOD detection performance on SVHN for models trained on
    CIFAR-10. Here, R-$r$ denotes LoRA rank $r$, and Ad-$m$ denotes
    $m$ learned adapters. EulerLoRA uses $T=2$ dynamics steps, while
    stochastic inference uses four Monte Carlo samples per adapter.
    Results use maximum softmax probability with in-distribution samples
    treated as positive and are reported as mean $\pm$ standard
    deviation over five seeds.}
    \label{tab:cifar10_svhn_ood}
    \resizebox{\columnwidth}{!}{
    \begin{tabular}{lccc}
        \toprule
        \textbf{Method}
        & \textbf{AUROC ($\uparrow$)}
        & \textbf{AUPRC ($\uparrow$)}
        & \textbf{FPR@95TPR ($\downarrow$)} \\
        \midrule

        LoRA (R-20, Ad-1)
        & $92.96 \pm 3.50$
        & $90.74 \pm 3.94$
        & $49.26 \pm 17.47$ \\

        LoRA-Ensemble (R-20, Ad-2)
        & $94.27 \pm 0.82$
        & $92.31 \pm 0.87$
        & $45.18 \pm 6.87$ \\

        \midrule

        \textbf{EulerLoRA} (R-20, Ad-1, Det.)
        & $93.81 \pm 1.63$
        & $91.25 \pm 2.17$
        & $44.72 \pm 8.44$ \\

        \textbf{EulerLoRA} (R-20, Ad-1, Stoch.)
        & $94.41 \pm 1.63$
        & $92.25 \pm 2.06$
        & $41.61 \pm 9.41$ \\

        \textbf{EulerLoRA} (R-20, Ad-2, Det.)
        & \secondbest{95.19}{0.28}
        & \best{93.44}{0.36}
        & \secondbest{39.09}{3.10} \\

        \textbf{EulerLoRA} (R-20, Ad-2, Stoch.)
        & \best{95.22}{0.24}
        & \secondbest{93.38}{0.27}
        & \best{37.42}{2.65} \\

        \bottomrule
    \end{tabular}
    }
\end{table}

\begin{table*}[t]
    \centering
    \caption{Performance comparison on HAM10000. Here, R-$r$ denotes
    LoRA rank $r$, and Ad-$m$ denotes $m$ learned adapters.
    EulerLoRA uses R-20, Ad-2, and $T=2$ dynamics steps.
    Stochastic inference uses four Monte Carlo samples per adapter,
    whereas deterministic inference uses all rank components.
    The controlled EulerLoRA results are reported as mean $\pm$
    standard deviation over five seeds.}
    \label{tab:ham10000_results}
    \resizebox{\linewidth}{!}{
    \begin{tabular}{lccccc}
        \toprule
        \textbf{Method}
        & \textbf{Accuracy ($\uparrow$)}
        & \textbf{F1 ($\uparrow$)}
        & \textbf{ECE ($\downarrow$)}
        & \textbf{NLL ($\downarrow$)}
        & \textbf{Brier ($\downarrow$)} \\
        \midrule

        Single Network
        & $84.1 \pm 0.3$
        & $71.4 \pm 0.7$
        & $0.139 \pm 0.004$
        & $1.138 \pm 0.040$
        & $0.291 \pm 0.009$ \\

        Single Net w/ LoRA
        & $83.2 \pm 0.7$
        & $70.7 \pm 1.3$
        & $0.085 \pm 0.004$
        & $0.569 \pm 0.027$
        & $0.256 \pm 0.011$ \\

        LRFormer
        & $74.3 \pm 1.9$
        & $52.1 \pm 3.2$
        & $0.053 \pm 0.022$
        & $0.737 \pm 0.014$
        & $0.354 \pm 0.011$ \\

        L2
        & $74.1 \pm 1.8$
        & $50.7 \pm 3.9$
        & $0.065 \pm 0.024$
        & $0.766 \pm 0.036$
        & $0.360 \pm 0.021$ \\

        MC Dropout
        & $83.7 \pm 0.4$
        & $71.0 \pm 0.9$
        & $0.099 \pm 0.007$
        & $0.631 \pm 0.023$
        & $0.270 \pm 0.009$ \\

        Snapshot Ensemble
        & $84.9 \pm 0.3$
        & $73.7 \pm 0.9$
        & $0.058 \pm 0.004$
        & $0.431 \pm 0.007$
        & $0.217 \pm 0.004$ \\

        Batch-Ensemble
        & $76.8 \pm 1.6$
        & $58.4 \pm 2.8$
        & $0.064 \pm 0.021$
        & $0.651 \pm 0.003$
        & $0.332 \pm 0.002$ \\

        Explicit Ensemble
        & $85.8 \pm 0.2$
        & $74.6 \pm 0.4$
        & $0.105 \pm 0.002$
        & $0.536 \pm 0.007$
        & $0.218 \pm 0.002$ \\

        LoRA-Ensemble (R-4, Ad-16)
        & $88.0 \pm 0.2$
        & $78.3 \pm 0.6$
        & \secondbest{0.037}{0.002}
        & \best{0.342}{0.003}
        & \secondbest{0.175}{0.002} \\

        \midrule

        \textbf{EulerLoRA} (R-20, Ad-2, Det.)
        & \best{89.6}{0.3}
        & \best{80.1}{0.9}
        & $0.040 \pm 0.003$
        & $0.385 \pm 0.009$
        & \best{0.164}{0.005} \\

        \textbf{EulerLoRA} (R-20, Ad-2, Stoch.)
        & \secondbest{89.1}{0.5}
        & \secondbest{79.4}{1.4}
        & \best{0.021}{0.003}
        & \secondbest{0.346}{0.009}
        & \best{0.164}{0.005} \\

        \bottomrule
    \end{tabular}
    }
\end{table*}

\begin{table*}[t]
    \centering
    \caption{Performance comparison on CIFAR-100. Here, R-$r$ denotes
    LoRA rank $r$, and Ad-$m$ denotes $m$ learned adapters.
    EulerLoRA uses R-20, Ad-2, and $T=2$ dynamics steps.
    Stochastic inference uses four Monte Carlo samples per adapter,
    whereas deterministic inference uses all rank components. 
    The deterministic and stochastic variants use the same trained
EulerLoRA checkpoints and differ only in the inference mode.
    The EulerLoRA results are reported as mean $\pm$ standard
    deviation over five seeds.}
    \label{tab:cifar100_results}
    \resizebox{\linewidth}{!}{
    \begin{tabular}{lccccc}
        \toprule
        \textbf{Method}
        & \textbf{Accuracy ($\uparrow$)}
        & \textbf{F1 ($\uparrow$)}
        & \textbf{ECE ($\downarrow$)}
        & \textbf{NLL ($\downarrow$)}
        & \textbf{Brier ($\downarrow$)} \\
        \midrule

        Single Network
        & $76.6 \pm 0.3$
        & $76.6 \pm 0.3$
        & $0.145 \pm 0.004$
        & $1.181 \pm 0.019$
        & $0.370 \pm 0.004$ \\

        Single Net w/ LoRA
        & $79.6 \pm 0.2$
        & $79.4 \pm 0.2$
        & \secondbest{0.014}{0.003}
        & $0.671 \pm 0.005$
        & $0.286 \pm 0.003$ \\

        MC Dropout
        & $77.1 \pm 0.5$
        & $77.2 \pm 0.4$
        & $0.055 \pm 0.002$
        & $1.138 \pm 0.014$
        & $0.336 \pm 0.005$ \\

        Last-layer Ensemble
        & $73.4 \pm 0.0$
        & $73.0 \pm 0.0$
        & $0.093 \pm 0.000$
        & $0.978 \pm 0.000$
        & $0.376 \pm 0.000$ \\

        Snapshot Ensemble
        & $77.0 \pm 0.1$
        & $77.2 \pm 0.2$
        & $0.123 \pm 0.002$
        & $4.416 \pm 0.046$
        & $1.614 \pm 0.007$ \\

        SNGP
        & $32.2 \pm 0.4$
        & $30.1 \pm 0.4$
        & $0.072 \pm 0.004$
        & $2.744 \pm 0.010$
        & $0.817 \pm 0.002$ \\

        ENN (EpiNet)
        & $79.7 \pm 0.2$
        & $79.7 \pm 0.2$
        & $0.128 \pm 0.003$
        & $1.016 \pm 0.015$
        & $0.323 \pm 0.005$ \\

        Batch-Ensemble
        & $68.8 \pm 0.1$
        & $68.5 \pm 0.1$
        & $0.102 \pm 0.002$
        & $1.093 \pm 0.002$
        & $0.437 \pm 0.001$ \\

        Explicit Ensemble
        & $79.8 \pm 0.1$
        & $79.8 \pm 0.2$
        & $0.100 \pm 0.001$
        & $0.745 \pm 0.003$
        & $0.284 \pm 0.002$ \\

        LoRA-Ensemble (R-8, Ad-16)
        & \secondbest{82.5}{0.1}
        & \best{82.5}{0.1}
        & $0.035 \pm 0.001$
        & \secondbest{0.587}{0.001}
        & \secondbest{0.253}{0.000} \\

        \midrule

        \textbf{EulerLoRA} (R-20, Ad-2, Stoch.)
        & \best{82.6}{0.2}
        & \secondbest{82.4}{0.2}
        & $0.058 \pm 0.002$
        & $0.589 \pm 0.003$
        & \secondbest{0.253}{0.001} \\

        \textbf{EulerLoRA} (R-20, Ad-2, Det.)
        & \best{82.6}{0.1}
        & \best{82.5}{0.1}
        & \best{0.013}{0.002}
        & \best{0.559}{0.003}
        & \best{0.244}{0.002} \\

        \bottomrule
    \end{tabular}
    }
\end{table*}

We evaluate EulerLoRA in terms of predictive performance, calibration,
probabilistic prediction, and out-of-distribution detection. We first
describe the experimental setup and then compare deterministic and
stochastic inference across the considered benchmarks.
\paragraph{Datasets.}
We evaluate EulerLoRA on CIFAR-10, CIFAR-100, and
HAM10000 \cite{tschandl2018ham10000} for in-distribution
classification, and use SVHN for OOD detection with CIFAR-10 as the
in-distribution dataset. For HAM10000, we follow the image-level
stratified $80/20$ split used by LoRA-Ensemble. Further dataset details
are provided in the supplementary material.

\paragraph{Backbone and LoRA configuration.}
All experiments use a pretrained ViT-B/32 with frozen backbone
parameters. LoRA modules are inserted into the query, key, value, and
output projections of each self-attention block. We use maximum rank
$r=20$, minimum active rank $K_{\min}=10$, and stochastic scale
$\sigma=1.0$. For the query, key, and value projections, one expectation-preserving
rank configuration is sampled per forward pass. The compensated
dynamics are applied only at the output projection, using $T=2$
internal steps with an independently sampled rank configuration at each
step. In all cases, selected rank components are scaled according to
Eq.~\eqref{eq:rank_mask}.

We evaluate one- and two-adapter settings on CIFAR-10, and use two
rank-$20$ adapters on CIFAR-100 and HAM10000. Stochastic inference
averages four Monte Carlo samples per adapter, while deterministic
inference activates all rank components in every projection, causing
the centered stochastic term at the output projection to vanish.
\paragraph{Trainable parameter count.}
The published LoRA-Ensemble with 16 rank-$8$ adapters contains
approximately $10$ million trainable adapter parameters. In contrast,
EulerLoRA with two rank-$20$ adapters contains approximately $3$
million trainable parameters. Thus, EulerLoRA reduces the trainable
adapter count by about $69\%$ while using substantially fewer
independently learned adapters

\paragraph{Training protocol.}
We follow the training configuration and random seeds used in
LoRA-Ensemble \cite{muehlematter2026loraensemble}. All models are trained
with cross-entropy loss and AdamW while keeping the pretrained ViT
parameters frozen. We train for 16 epochs on CIFAR-10 and CIFAR-100,
and for 65 epochs on HAM10000. All controlled rank-$20$ experiments
are repeated over five seeds, and we report the mean and standard
deviation across runs. Stochastic and deterministic results are
obtained from the same trained EulerLoRA checkpoints and differ only
in the inference mode. Additional implementation details, complete
hyperparameter settings, random-seed configuration, and computational
resources are provided in the supplementary material.

\paragraph{Baselines.}
We compare EulerLoRA against the baselines reported by
LoRA-Ensemble \cite{muehlematter2026loraensemble}; unless stated otherwise,
the corresponding results are taken directly from that work rather than
reimplemented in our study. These baselines include a \emph{Single
Network} and a \emph{Single Net with LoRA}, which use one pretrained
ViT without and with low-rank adaptation, respectively. The reported
\emph{Explicit Ensemble} independently fine-tunes multiple models and
serves as a strong but computationally expensive reference
\cite{lakshminarayanan2017simple}. The comparison also includes MC
Dropout \cite{gal2016dropout}, Snapshot Ensemble
\cite{huang2017snapshot}, BatchEnsemble
\cite{wen2020batchensemble}, and a Last-layer Ensemble based on a shared
feature extractor with multiple predictors. For CIFAR-100, we additionally report the
published results for EpiNet \cite{osband2023epistemic} and SNGP
\cite{liu2020simple}. For HAM10000, we include the reported L2
self-attention and LRFormer results
\cite{kim2021lipschitz,ye2023lrformer}. Finally, LoRA-Ensemble shares a
frozen pretrained backbone while learning an independent set of LoRA
factors for each ensemble member
\cite{muehlematter2026loraensemble}. Our own reproduced LoRA and
LoRA-Ensemble controls are explicitly marked as such in the tables.

\paragraph{Evaluation metrics.}
For in-distribution classification, we report accuracy, macro-F1,
expected calibration error (ECE), negative log-likelihood (NLL), and
the multiclass Brier score. For OOD detection, we use maximum softmax
probability (MSP), treating in-distribution samples as the positive
class, and report AUROC, AUPRC, and FPR@95TPR. Higher accuracy, macro-F1,
AUROC, and AUPRC are better, whereas lower ECE, NLL, Brier score, and
FPR@95TPR are better. Formal definitions are provided in the
supplementary material.

\subsection{Predictive Performance and Calibration}

Table~\ref{tab:cifar10_results} reports the results on CIFAR-10.
EulerLoRA consistently improves over the matched deterministic LoRA
baseline. With one rank-$20$ adapter, stochastic inference increases
accuracy from $94.66\%$ to $95.44\%$ and reduces NLL from $0.173$ to
$0.135$. The Brier score also decreases from $0.083$ to $0.069$,
showing that the improvement is not limited to classification accuracy
but extends to the quality of the predictive probabilities. The
deterministic evaluation of the same one-adapter EulerLoRA model is
weaker than its stochastic counterpart, indicating that averaging
rank-sampled trajectories is particularly beneficial when only one
adapter is available.
Increasing the number of adapters from one to two produces a further
improvement. The two-adapter stochastic model reaches $95.92\%$
accuracy, an NLL of $0.126$, and a Brier score of $0.064$. Deterministic
inference provides the strongest overall CIFAR-10 performance, achieving
$96.10\%$ accuracy, an NLL of $0.116$, and a Brier score of $0.059$.
It also attains a low ECE of $0.004$. Thus, stochastic training does
not require stochastic inference to remain useful: retaining all rank
components at test time can yield a strong deterministic predictor,
while Monte Carlo rank sampling remains available when predictive
diversity is required.

Compared with the reproduced two-adapter LoRA-Ensemble, deterministic
EulerLoRA improves accuracy by $0.67$ percentage points and reduces NLL
from $0.138$ to $0.116$. The stochastic variant also improves accuracy,
NLL, and Brier score, although its ECE of $0.016$ is worse than the
$0.007$ obtained by the reproduced LoRA-Ensemble. 

EulerLoRA is also competitive with the substantially larger published
LoRA-Ensemble containing 16 rank-$8$ adapters. The two-adapter
deterministic configuration improves accuracy from $95.9\%$ to
$96.10\%$, NLL from $0.128$ to $0.116$, and Brier score from
$0.064$ to $0.059$, while obtaining a comparable ECE. The stochastic
configuration approximately matches the published ensemble in accuracy,
NLL, and Brier score. Importantly, the published LoRA-Ensemble uses
approximately $10$ million trainable adapter parameters, whereas
EulerLoRA uses approximately $3$ million. These results show that
EulerLoRA recovers the predictive benefits of a much larger adapter
ensemble with about $69\%$ fewer trainable parameters.

Table~\ref{tab:ham10000_results} shows a similar advantage on HAM10000.
Deterministic EulerLoRA achieves the highest accuracy and macro-F1,
reaching $89.6\%$ and $80.1\%$, respectively. This exceeds the
16-adapter LoRA-Ensemble by $1.6$ percentage points in accuracy and
$1.8$ points in macro-F1. Both EulerLoRA inference modes also obtain
the best Brier score of $0.164$, compared with $0.175$ for
LoRA-Ensemble.
The two inference modes again provide complementary behavior.
Deterministic inference gives the best accuracy and macro-F1, whereas
stochastic inference yields the best ECE of $0.021$, reducing the ECE
of LoRA-Ensemble from $0.037$. Its NLL of $0.346$ is also close to the
LoRA-Ensemble value of $0.342$. Therefore, on the imbalanced
HAM10000 dataset, averaging stochastic rank trajectories improves
calibration, while deterministic inference preserves the strongest
classification performance.

On CIFAR-100, deterministic EulerLoRA achieves $82.6\%$ accuracy
and matches the best macro-F1 of $82.5\%$. It also provides the
strongest calibration and proper scoring performance, reducing ECE
from $0.035$ to $0.013$, NLL from $0.587$ to $0.559$, and the Brier
score from $0.253$ to $0.244$ relative to the 16-adapter
LoRA-Ensemble. Stochastic inference retains comparable accuracy and
Brier performance, although its ECE is higher. These results again
show that stochastic training can improve the all-ranks-active
deterministic predictor.
\subsection{Out-of-Distribution Detection}

Table~\ref{tab:cifar10_svhn_ood} evaluates models trained on CIFAR-10 using
SVHN as out-of-distribution data. Standard one-adapter LoRA obtains an
AUROC of $92.96\%$ and an FPR@95TPR of $49.26\%$. Introducing
rank-driven dynamics improves both metrics even with a single adapter.
Deterministic EulerLoRA increases AUROC to $93.81\%$ and reduces
FPR@95TPR to $44.72\%$, while stochastic inference further improves
them to $94.41\%$ and $41.61\%$, respectively.

Using two adapters produces a larger improvement. The deterministic
EulerLoRA model achieves an AUROC of $95.19\%$, an AUPRC of
$93.44\%$, and an FPR@95TPR of $39.09\%$. Stochastic inference
provides the best overall OOD separation, with an AUROC of $95.22\%$
and an FPR@95TPR of $37.42\%$. Relative to the reproduced two-adapter
LoRA-Ensemble, this lowers FPR@95TPR by $7.76$ percentage points,
while increasing AUROC by approximately one percentage point.

The difference between deterministic and stochastic inference is small
for AUROC and AUPRC but more pronounced for FPR@95TPR. This suggests
that trajectory averaging mainly improves the difficult operating
region in which $95\%$ of the in-distribution samples must be retained.
The result is consistent with the intended role of stochastic
rank-driven dynamics: different sampled rank configurations provide
additional predictive variation without requiring a separate learned
adapter for each Monte Carlo realization.

\paragraph{Overall analysis.}
Across the three datasets, deterministic and stochastic inference
provide complementary behavior. Deterministic inference generally
achieves the strongest classification accuracy and proper scoring
performance, showing that stochastic rank-driven training also benefits
the all-ranks-active predictor. Stochastic inference is particularly
effective for calibration on HAM10000 and for OOD detection on SVHN,
where it achieves the lowest ECE and FPR@95TPR, respectively.
Importantly, these results are obtained using only two rank-$20$
adapters, compared with 16 independently learned adapters in the
published LoRA-Ensemble baselines.
\section{Conclusion}

We introduced EulerLoRA, a stochastic extension of LoRA that evolves
the rank-one components of the low-rank update through compensated
finite-step dynamics. The formulation preserves the deterministic LoRA
transformation in expectation while generating multiple predictive
trajectories from shared adapter parameters. Across CIFAR-10, CIFAR-100, HAM10000, and SVHN OOD detection,
EulerLoRA achieves competitive or improved accuracy, calibration, and
uncertainty estimation compared with LoRA-Ensemble, while using only
two rank-$20$ adapters instead of up to 16 independently learned
adapters. Deterministic inference generally provides the strongest
predictive performance, whereas stochastic inference offers
complementary gains in calibration and OOD detection. These results
show that useful predictive diversity can be obtained with about
$69\%$ fewer trainable adapter parameters.
\bibliography{aaai2027}
\clearpage
\section*{Supplementary Material Overview}

This document provides additional dataset details, implementation and
training settings, random-seed configuration, computational resources,
formal definitions of the evaluation metrics, and the trainable-parameter
calculation supporting the results reported in the main paper.

\section{Dataset Details}
\label{sec:datasets}

\paragraph{CIFAR-10 and CIFAR-100.}
CIFAR-10 and CIFAR-100 contain $50{,}000$ training images and $10{,}000$
test images of size $32\times32$ pixels, with 10 and 100 classes,
respectively \citep{krizhevsky2009learning}. We use the standard training and
test partitions. CIFAR-10 is used for in-distribution classification and as
the in-distribution dataset in the SVHN OOD experiment. CIFAR-100 is used for
in-distribution classification.

\paragraph{HAM10000.}
HAM10000 contains $10{,}015$ dermatoscopic images from seven diagnostic
categories \citep{tschandl2018ham10000}. We follow the image-level split used
by LoRA-Ensemble \citep{muehlematter2026loraensemble}: the metadata are divided
into a stratified $80/20$ train/test split using the numeric class label and a
fixed split seed of 42. 

\paragraph{SVHN for OOD detection.}
Models trained on CIFAR-10 are evaluated using the standard CIFAR-10 test set
as in-distribution data and the SVHN test split as out-of-distribution data
\citep{netzer2011reading}. The SVHN test split contains $26{,}032$ images. No
SVHN image is used during training or hyperparameter selection.

\begin{table*}[t]
\centering
\caption{Dataset summary. HAM10000 uses an image-level stratified $80/20$
split. SVHN is used only for OOD evaluation.}
\begin{tabular}{lrrrrl}
\toprule
Dataset & Classes & Train & Test & Native size & Role \\
\midrule
CIFAR-10  & 10  & 50,000 & 10,000 & $32\times32$   & ID classification and OOD ID \\
CIFAR-100 & 100 & 50,000 & 10,000 & $32\times32$   & ID classification \\
HAM10000  & 7   & 8,012  & 2,003  & $450\times600$ & ID classification \\
SVHN      & 10  & --     & 26,032 & $32\times32$   & OOD test set \\
\bottomrule
\end{tabular}

\label{tab:dataset_summary}
\end{table*}

\subsection{Image Preprocessing and Augmentation}

Input images are converted to floating-point tensors and rescaled by
a factor of $1/255$. During training, we additionally apply independent
random horizontal and vertical flips, each with probability $0.5$, and
a random rotation with an angle sampled uniformly between $0^\circ$
and $180^\circ$. All images are resized to $224\times224$ before being
passed to ViT-B/32. During evaluation, only the $1/255$ rescaling and
resizing are applied.

\section{Architecture and EulerLoRA Implementation}
\label{sec:implementation}

\subsection{Backbone and Trainable Modules}
All experiments use an ImageNet-pretrained ViT-B/32
\citep{dosovitskiy2021image}. The backbone has 12 transformer blocks, a hidden
dimension of 768, and 12 attention heads. The pretrained backbone parameters
are frozen. A separate classification head is learned for each adapter, and
LoRA factors \citep{hu2022lora} are inserted into the query, key, value, and
output projections of every self-attention block. For projection
$P\in\{Q,K,V,O\}$,
\begin{equation}
    P(X)=W_P^0X+B_PA_PX,
\end{equation}
where $W_P^0$ is frozen, $A_P\in\R^{r\times768}$, and
$B_P\in\R^{768\times r}$ are trainable.

The controlled experiments use maximum rank $r=20$, minimum active rank
$K_{\min}=10$, and two independently parameterized adapters, except for the
additional one-adapter CIFAR-10 configuration. Each adapter has its own LoRA
factors and classification head; the pretrained ViT backbone is shared.

\subsection{Random Active-Rank Sampling}
At every stochastic sampling event, the active rank is drawn uniformly:
\begin{equation}
    K\sim\mathcal{U}\{K_{\min},K_{\min}+1,\ldots,r\}.
\end{equation}
Conditioned on $K$, a subset $S\subseteq\{1,\ldots,r\}$ of cardinality $K$
is sampled uniformly without replacement. Component $k$ receives coefficient
\begin{equation}
    M_k=\frac{r}{K}\1\{k\in S\}.
    \label{eq:mask}
\end{equation}
Since every component is selected with probability $K/r$,
\begin{equation}
    \E[M_k\mid K]=\frac{r}{K}\frac{K}{r}=1.
\end{equation}

\subsection{Sampling at the Query, Key, and Value Projections}
For $P\in\{Q,K,V\}$, one independent rank configuration is sampled per
projection and stochastic forward pass. These projections use a single
rank-sampled LoRA update and do not use the $T$-step dynamics:
\begin{equation}
 P_M(X)=W_P^0X+\sum_{k=1}^{r}M_{P,k}b_{P,k}a_{P,k}^{\top}X.
 \label{eq:qkv_sample}
\end{equation}
The three projection modules sample separately. Masks are therefore
independent across $Q$, $K$, and $V$, across stochastic forward passes, and,
in the multi-adapter setting, across adapters. The linear projection is
expectation-preserving:
\begin{equation}
 \E[P_M(X)\mid X,K]=W_P^0X+B_PA_PX.
 \label{eq:qkv_mean}
\end{equation}
Because self-attention contains nonlinear operations, including the
softmax, this projection-level expectation result does not imply that
the output of the complete stochastic attention block is an unbiased
estimator of the corresponding deterministic attention block.

\subsection{Compensated Dynamics at the Output Projection}
Only the output projection uses multiple internal steps. Let $C$ be the
attention representation entering this projection and let $\Delta t=1/T$.
The reported experiments use $T=2$, stochastic scale $\sigma=1$, and an
independently resampled rank configuration at each internal step. Starting from $Y_0=0$, the update is,
\begin{equation}
\begin{aligned}
Y_{s+1}
={}&Y_s
+\Delta t\left(W_O^0 C + BAC\right)\\
&+\sigma\sqrt{\Delta t}
\sum_{k=1}^{r}(M_{s,k}-1)b_k a_k^\top C \\
s=0,\ldots,T-1.
\end{aligned}
\end{equation}
The attention representation $C$ remains fixed during the internal steps; the
frozen output projection is not repeatedly composed with itself.


\subsection{Initialization and Inference Modes}
For every LoRA pair, $A$ is initialized with Xavier-uniform initialization
using gain 10, while $B$ is initialized to zero. Thus the initial LoRA
correction $BA$ is zero.

Deterministic and stochastic results use identical trained checkpoints. In
stochastic inference, rank sampling remains active and four Monte Carlo
samples are drawn per adapter. In deterministic inference, rank sampling is
disabled and every learned rank component is active, so $K=r$ and $M_k=1$.
The centered output-projection term then vanishes. No parameters are retrained
or modified between inference modes.

For $A$ adapters and $S$ stochastic samples per adapter, the model produces
$AS$ predictive trajectories. The final distribution is the arithmetic mean
of the trajectory-wise softmax probabilities:
\begin{equation}
 \bar p(y=c\mid x)=\frac{1}{AS}
 \sum_{a=1}^{A}\sum_{s=1}^{S}
 \softmax\bigl(z_{a,s}(x)\bigr)_c.
 \label{eq:prob_average}
\end{equation}


\section{Training Configuration and Reproducibility}
\label{sec:training}

\subsection{Optimization}
The pretrained ViT parameters remain frozen. We optimize the LoRA factors and
classification heads using cross-entropy loss and AdamW
\citep{loshchilov2019decoupled}. The learning rate is linearly warmed from
zero to $10^{-4}$ during the first 500 optimization steps and then follows a
cosine decay schedule. Gradients are clipped to maximum norm 1.0. Automatic
mixed precision and early stopping are disabled. CIFAR-10 and CIFAR-100 use
uniform class weights. HAM10000 uses class-balanced weights based on the
effective number of samples \citep{cui2019classbalanced}, with
$\beta=0.9991$.

Table~\ref{tab:training_config} lists the exact dataset-specific settings. The
weight-decay difference between CIFAR-100 and the other two datasets is
retained exactly as used in the reported runs.

\begin{table*}[t]
\centering
\setlength{\tabcolsep}{5pt}
\begin{tabular}{lccc}
\toprule
Setting & CIFAR-10 & CIFAR-100 & HAM10000 \\
\midrule
Epochs / maximum steps & 16 / 25,008 & 16 / 25,008 & 65 / 16,315 \\
Training / evaluation batch size & 32 / 128 & 32 / 128 & 32 / 128 \\
Optimizer and learning rate & AdamW, $10^{-4}$ & AdamW, $10^{-4}$ & AdamW, $10^{-4}$ \\
Adam $(\beta_1,\beta_2)$ & $(0.9,0.999)$ & $(0.9,0.999)$ & $(0.9,0.999)$ \\
Weight decay & 0.01 & 0.01 & 0.01 \\
Warm-up / schedule & 500 / cosine & 500 / cosine & 500 / cosine \\
Gradient clipping & 1.0 & 1.0 & 1.0 \\
Loss and class weights & CE, uniform & CE, uniform & weighted CE, $\beta=0.9991$ \\
Input resolution & $224\times224$ & $224\times224$ & $224\times224$ \\
Training transforms & flip, rotate, rescale & flip, rotate, rescale & flip, rotate, rescale \\
Evaluation transforms & rescale & rescale & rescale \\
\bottomrule
\end{tabular}
\caption{Dataset-specific training and evaluation configuration.}
\label{tab:training_config}
\end{table*}

\subsection{Random Seeds and Reporting Protocol}
All controlled rank-20 experiments are repeated using the five seeds
\begin{equation}
    \{0,\ 42,\ 1206,\ 2205,\ 25008\}.
\end{equation}
The selected seed is passed to the repository's global seeding utility before
model construction and training. For every seed, deterministic and stochastic
inference are evaluated from the same checkpoint. Main-paper tables report
the mean and standard deviation over these five runs.

\subsection{Computational Resources}
The reported runs were executed on an NVIDIA A100-SXM4 GPU with 40~GB memory.
Each experiment used two data-loader workers. The adapter ensemble was
processed on one GPU and automatic mixed precision was disabled. The
implementation uses PyTorch, builds on the official LoRA-Ensemble codebase,
and loads pretrained ViT weights through torchvision.

\section{Evaluation Metrics}
\label{sec:metrics}

Let $N$ be the number of examples, $C$ the number of classes,
$y_i\in\{1,\ldots,C\}$ the true label, and $\bar p_{i,c}$ the mean predictive
probability from Equation~\eqref{eq:prob_average}. The predicted label and
confidence are
\begin{equation}
    \widehat y_i=\arg\max_c\bar p_{i,c},
    \qquad q_i=\max_c\bar p_{i,c}.
\end{equation}

\subsection{Accuracy and Macro-F1}
Accuracy is
\begin{equation}
    \operatorname{Acc}=\frac{1}{N}\sum_{i=1}^{N}\1\{\widehat y_i=y_i\}.
\end{equation}
For class $c$, let $\operatorname{Prec}_c$ and $\operatorname{Rec}_c$ denote
precision and recall. Macro-F1 is
\begin{equation}
    \operatorname{MacroF1}=\frac{1}{C}\sum_{c=1}^{C}
    \frac{2\operatorname{Prec}_c\operatorname{Rec}_c}
    {\operatorname{Prec}_c+\operatorname{Rec}_c}.
\end{equation}

\subsection{Expected Calibration Error}
We use fixed-width ECE \citep{guo2017calibration} with $M=10$ equally spaced
confidence bins over $[0,1]$. For examples $B_m$ in bin $m$, define
\begin{equation}
 \operatorname{acc}(B_m)=\frac{1}{|B_m|}
 \sum_{i\in B_m}\1\{\widehat y_i=y_i\},
\end{equation}
\begin{equation}
 \operatorname{conf}(B_m)=\frac{1}{|B_m|}\sum_{i\in B_m}q_i.
\end{equation}
Then
\begin{equation}
 \operatorname{ECE}=\sum_{m=1}^{M}\frac{|B_m|}{N}
 \left|\operatorname{acc}(B_m)-\operatorname{conf}(B_m)\right|.
\end{equation}

\subsection{Negative Log-Likelihood and Brier Score}
The multiclass negative log-likelihood is
\begin{equation}
    \operatorname{NLL}=-\frac{1}{N}\sum_{i=1}^{N}\log\bar p_{i,y_i}.
\end{equation}
The multiclass Brier score \citep{brier1950verification} is
\begin{equation}
    \operatorname{Brier}=\frac{1}{N}\sum_{i=1}^{N}\sum_{c=1}^{C}
    \left(\bar p_{i,c}-\1\{y_i=c\}\right)^2.
\end{equation}
Lower NLL and Brier score indicate better probabilistic predictions.

\subsection{OOD Score and Positive-Class Convention}
We use maximum softmax probability (MSP) \citep{hendrycks2017baseline}:
\begin{equation}
    s_{\mathrm{MSP}}(x)=\max_c \bar p(y=c\mid x).
\end{equation}
Higher scores indicate that an example is more likely to be in-distribution.
ID examples are assigned the positive label and OOD examples the negative
label for all reported OOD metrics.

\subsection{AUROC and AUPRC}
The receiver operating characteristic curve plots
\begin{equation}
    \operatorname{TPR}(\tau)=\frac{\operatorname{TP}(\tau)}
    {\operatorname{TP}(\tau)+\operatorname{FN}(\tau)}
\end{equation}
against
\begin{equation}
    \operatorname{FPR}(\tau)=\frac{\operatorname{FP}(\tau)}
    {\operatorname{FP}(\tau)+\operatorname{TN}(\tau)}
\end{equation}
as threshold $\tau$ varies. AUROC is the area under this curve. AUPRC is the
area under the precision-recall curve, where
\begin{equation}
    \operatorname{Precision}(\tau)=\frac{\operatorname{TP}(\tau)}
    {\operatorname{TP}(\tau)+\operatorname{FP}(\tau)},
\end{equation}
\begin{equation}
    \operatorname{Recall}(\tau)=\operatorname{TPR}(\tau).
\end{equation}
Because ID is positive, AUROC and AUPRC quantify the ability to rank ID
examples above OOD examples.

\subsection{FPR at 95\% TPR}
FPR@95TPR is the false-positive rate at an operating point retaining at least
95\% of ID examples. Among finite-sample ROC thresholds satisfying
$\operatorname{TPR}(\tau)\geq0.95$, we report
\begin{equation}
    \operatorname{FPR@95TPR}
    =\min_{\tau:\operatorname{TPR}(\tau)\geq0.95}
      \operatorname{FPR}(\tau).
\end{equation}
A false positive is an OOD example whose MSP exceeds the ID-acceptance
threshold and is therefore accepted as in-distribution. Lower values are
better.

\section{Trainable Parameter Calculation}
\label{sec:params}

\begin{table}[t]
\centering
\small
\setlength{\tabcolsep}{2.5pt}
\begin{tabular}{lrrr}
\toprule
Dataset & LoRA & Heads & Total \\
\midrule
CIFAR-10  & 2,949,120 & 15,380  & 2,964,500 \\
CIFAR-100 & 2,949,120 & 153,800 & 3,102,920 \\
HAM10000  & 2,949,120 & 10,766  & 2,959,886 \\
\bottomrule
\end{tabular}
\caption{Trainable parameters for two rank-20 EulerLoRA adapters. The
pretrained backbone is frozen and excluded.}
\label{tab:param_counts}
\end{table}

For one square $768\times768$ attention projection, a rank-$r$ LoRA pair
contains
\begin{equation}
    768r+r768=1536r
\end{equation}
parameters. LoRA is applied to four projections in each of 12 blocks, so one
adapter contains
\begin{equation}
    N_{\mathrm{adapter}}(r)=12\times4\times1536r=73{,}728r.
\end{equation}
Therefore,
\begin{align}
N_{\method{}}(r=20,A=2)
    &=73{,}728\times20\times2 \\
    &=2{,}949{,}120,
\end{align}
whereas the rank-8, 16-adapter LoRA-Ensemble contains
\begin{align}
N_{\mathrm{LE}}(r=8,A=16)
    &=73{,}728\times8\times16 \\
    &=9{,}437{,}184.
\end{align}
The reduction in trainable adapter parameters is
\begin{equation}
    1-\frac{2{,}949{,}120}{9{,}437{,}184}=0.6875,
\end{equation}
or $68.75\%$. These exact values correspond to the approximately 3-million
versus 10-million adapter counts in the main paper.

The comparison above excludes task heads from both methods. A two-adapter
EulerLoRA model additionally contains one linear $768\rightarrow C$ head per
adapter. Table~\ref{tab:param_counts} gives the corresponding total trainable
counts. The CIFAR-100 total, $3{,}102{,}920$, matches the value emitted by the
training code.

\section{Reproducibility Summary}
The principal experimental configuration is reproduced by the following
steps:
\begin{enumerate}
    \item Load an ImageNet-pretrained ViT-B/32 and replace its task head.
    \item Freeze the pretrained ViT and insert rank-20 LoRA factors into
    $Q,K,V,O$ of all 12 attention blocks.
    \item Create one or two independently parameterized adapters and heads.
    \item At every stochastic event, draw $K$ uniformly from
    $\{10,\ldots,20\}$, select $K$ components without replacement, and scale
    the selected components by $20/K$.
    \item Use one independent sample at each of $Q,K,V$ and two compensated
    steps at $O$, resampling the output mask at each step.
    \item Train with four stochastic samples per adapter and the settings in
    Table~\ref{tab:training_config} for each of the five seeds.
    \item For stochastic evaluation, average four softmax probability vectors
    per adapter. For deterministic evaluation, use the same checkpoint with
    all rank components active.
    \item Compute the metrics in Section~\ref{sec:metrics} and aggregate means
    and standard deviations across the five runs.
\end{enumerate}

\end{document}